\documentclass{article}

\usepackage{arxiv}

\usepackage[utf8]{inputenc} 
\usepackage[T1]{fontenc}    
\usepackage[bookmarks=false]{hyperref}       
\usepackage{url}            
\usepackage{booktabs}       
\usepackage{amsfonts}       
\usepackage{nicefrac}       
\usepackage{microtype}      
\usepackage{lipsum}		
\usepackage{graphicx}
\usepackage{amsmath, amssymb}
\usepackage{xspace}
\usepackage{textcomp}
\newcommand{\mz}{\ensuremath{m/z}\xspace}
\newcommand{\norm}[1]{\ensuremath{\left\lVert{#1}\right\rVert_2}}

\title{SoRC - Evaluation of Computational Molecular Co-Localization Analysis in Mass Spectrometry Images}

\author{ \href{https://orcid.org/0000-0002-3935-7598}{\includegraphics[scale=0.06]{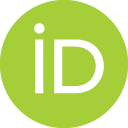}\hspace{1mm}Karsten Wüllems} \\
	International DFG Research Training Group GRK 1906\\
	Biodata Mining Group, Faculty of Technology\\	
	Bielefeld University\\
	Bielefeld, Germany \\
	\texttt{wuellems@cebitec.uni-bielefeld.de} \\
	\And
	Tim W. Nattkemper \\
	Biodata Mining Group, Faculty of Technology\\	
	Bielefeld University\\
	Bielefeld, Germany \\
	\texttt{tim.nattkemper@uni-bielefeld.de} \\
}

\renewcommand{\shorttitle}{SoRC}

\hypersetup{
pdftitle={SoRC - Evaluation of Computational Molecular Co-Localization Analysis in Mass Spectrometry Images},
pdfsubject={stat.ME, stat.AP, stat.CO},
pdfauthor={Karsten Wüllems, Tim W. Nattkemper},
pdfkeywords={Mass Spectrometry Imaging, Clustering, Similarity Functions, Image Processing, Co-Localization Analysis},
}

\begin{document}
\maketitle

\begin{abstract}
\textbf{Background:} The computational analysis of Mass Spectrometry Imaging (MSI) data aims at the identification of interesting mass co-localizations and the visualization of their lateral distribution in the sample, usually a tissue cross section. But as the morphological structure of tissues and the different kinds of mass co-localization naturally show a huge diversity, the selection and tuning of the computational method is a time-consuming effort. In this work we address the special problem of computationally grouping mass channel images according to their similarities in their lateral distribution patterns. Such an analysis is driven by the idea, that groups of molecules that feature a similar distribution pattern may have a functional relation. But the selection of the similarity function and other parameters is often done by a time-consuming and unsatsifactory trial and error. We propose a new flexible workflow scheme called SoRC (\textbf{s}um \textbf{o}f \textbf{r}anked \textbf{c}luster indices) for automating this tuning step and making it much more efficient.  

\noindent \textbf{Results:}
We test SoRC using three different data sets acquired from the lab for three different kinds of samples (barley seed, mouse bladder tissue, human PXE skin). We show, that SORC can be applied to score and visualize the results obtained with the applied methods in short time without too much effort. In our application example, the SoRC results for the three data sets reveal that a) some well-known similarity functions are suited to achieve good results for all three data sets and b) for the MSI data featuring a higher degree of irregularity improved results can be achieved by applying non-standard similarity functions. 

\noindent \textbf{Conclusion:}
The SoRC scores computed with our approach indicate that an automated testing and scoring of different methods for mass channel image grouping can improve the final outcome of a study by finally selecting the methods of the highest scores.
\end{abstract}


\section{Introduction}
Mass spectrometry imaging (MSI) is a well established and widely used technique to identify the spatial distributions of various molecules within biological samples. The extension of this technology into ever new domains leads to a constant increase in the amount of data and in the diverity of biological samples investigated with this technology. The high dimension of MSI analysis of any MSI data is a complex problem as it combines two domains of information. On the one hand, there is the information provided by many mass spectra. On the other hand, there is the spatial information provided by molecular distributions. In addition, both are influenced by the morphology of the biological sample.

A computational approach to MSI analysis can either target the grouping (or clustering) of the of pixels and their mass spectra \cite{mccombie2005spatial,deininger2008maldi,alexandrov2010spatial,kolling2012whide} or the grouping of masses and their associated lateral molecular distribution patterns \cite{alexandrov2013analysis,wijetunge2014unsupervised,wullems2019detection}.
The first approach often pursues the objective of grouping pixels into clusters and the final clustering outcome is visualized as a so-called segmentation map \cite{alexandrov2010spatial,kolling2012whide}. These segmentation maps use colors to present the location of each cluster in the MSI visualization. The second approach pursues grouping of similar lateral molecular distribution patterns into clusters. These molecular distribution patterns are often called mass channel images and are visualized as grey value or heat map visualizations.
The idea behind this approach is that co-localization of molecules can hint to molecular interactions, which in turn can be an indication for a relationship within a molecular network. An important requirement for the clustering of mass channel images is the non-trivial problem of the definition of a function to map pairs of mass channel images to similarity or dissimilarity values. 

Since the analysis of molecular distributions by mass spectrometry imaging is mainly focused around the analysis of mass
channel images, there is a close relation to the field of image processing.
The field of image processing has proposed several functions to quantify the similarity of two images. It has been shown that the quantification of similarity between two images can not be defined in a unique way, because visual similarity is often context dependent. There are examples, like natural scenes \cite{simoncelli2001natural,srivastava2003advances,sheikh2005information} or face images \cite{abaza2012quality}, where it has already been shown that some similarity functions perform significantly better than others. Transferring the problem of defining pairwise similarity to the field of MSI we realize the same kind of context dependency. First, there is a biological context, which can be described by the biological and biochemical characteristics of the sample, such as the morphology, molecular networks, biochemical reactions and functions etc. Second, there is the technical context, which is given by the measurement conditions, e.g. the measuring instrument, parameter settings or the matrix protocol applied. This multitude of contexts raises the question if some algorithms are more suitable than others to assess the similarity between two mass channel images in one particular combination of biological and technical contexts. This question becomes even more difficult to answer considering the fact that the grouping of mass images is usually achieved using a pipeline of inter-related consecutive computational steps. These steps include for instance methods for denoising, image/signal enhancement, image transformations and clustering methods. In this manuscript, one particular pipeline of parameterized algorithmic modules as a pipeline setup.

Many previous works on MSI data and grouping mass channel images preferably used quite basic functions for assessing the similarity of mass channel images, such as the \emph{Pearson Correlation Coefficient} \cite{mather1992cluster}, the \emph{Cosine Similarity} \cite{mather1992cluster} or the \emph{Structural Similarity Index} \cite{wang2004image}. But most of these works do not provide any motivation for the choice of their function. To the best of our knowledge, there is only one comparative study on different similarity measures for co-localization, which also investigates the influence of some basic pre-processing methods \cite{ovchinnikova2020colocml}. However, the number of methods is considerably smaller compared to our study, the pipeline setup plays a much smaller role and, because of the individual focus of this work, no systematic evaluation procedure is presented. 

In this manuscript three biological samples are used for testing our approach in three different biological contexts. We will investigate the influence of the pipeline setup, with a focus on various similarity measures, which are mostly derived from the field of image processing. In addition, we will propose a methodology called SoRC (\textbf{s}imilarity \textbf{o}f \textbf{r}anked \textbf{c}luster indices). SoRC is a workflow scheme to evaluate the suitability of various pipeline setups. Part of the workflow is the computation of a score for each context, called SoRC score. The score is computed by utilizing two statistical measures that estimate the quality of clusters, called cluster indices. The SoRC scores will be visualized to suggest suitable combinations of methods for a given biological sample and a defined clustering method. This way it will support the user to design and evaluate a good analysis pipeline to cluster the mass channel images of a given biological sample.

\section{Materials and Methods}

\subsection{Data Set Definition}
This section briefly introduces the formal description of an MSI data set as used in this manuscript. One mass spectrometry image data set is defined as a multivariate image $\mathcal{I}$ that consists of $H \in \mathbb{N}^+$ rows and $W \in \mathbb{N}^+$ columns of pixels $\mathcal{I}_{i,j}$, with $i \in \{0, \ldots , H-1\}$ and  $j \in \{0, \ldots , W-1\}$. The subset of all spectral positions is defined as $\rho=\{(i,j) \ | \ i \in \{0, \ldots , H-1\}, j \in \{0, \ldots , W-1\}\}$. This set describes all positions where the MSI instrument applied a measurement, i.e. a mass spectrum has been recorded. The total number of all spectral positions for $i$ and $j$ is defined by $H$ and $W$. Each spectral pixel consists of $Z \in \mathbb{N^+}$ measured intensity values, each of which corresponds to a specific mass channel. These mass channels are indexed by $z \in \{0, \ldots , Z-1\}$ and are also named mass-to-charge ratio values (\mz-values), each of which corresponds to the molecular mass. At each spectral position with $(i,j) \in \rho$, the term $\mathcal{I}_{i,j,z}$ represents the intensity value of mass channel $z$. For all other positions $(i,j)\notin \rho$ we set $\mathcal{I}_{i,j,z} = 0 \ \forall \ z$. 

It is known that MSI technology suffers from a large variation of different intensity value ranges between individual mass channel images. To compensate for this, each mass channel image was initially linearly scaled into a range of $[0,1]$. However, due to write and read operations with the viridis colormap the range shifted into $[0.11765, 0.84314]$.

\subsection{Mass Channel Image Features and Representations}
We define the following feature maps and image representations in order to encode pixel-related mass signal characteristics.
\paragraph{Gradient vector image:} To integrate the local signal change and its direction (i.e. the signal value gradient) within a mass channel image we compute a gradient vector image $\mathcal{I}^{\nabla}_z$ for each mass.

\begin{minipage}[t]{0.4\textwidth}
\begin{equation}
 \mathcal{I}^{\nabla}_z = \nabla_{i,j} \mathcal{I}_z = \bigg(\overrightarrow{{\scriptstyle \nabla_i \mathcal{I}_{z}} \atop {\scriptstyle \nabla_j \mathcal{I}_{z}}}\bigg) \label{eq:vectormap} \\
\end{equation}
\end{minipage}
\hfill
\begin{minipage}[t]{0.25\textwidth}
\begin{equation}
 \nabla_i \mathcal{I}_{z} = \frac{\partial \mathcal{I}_{z}}{\partial i}
\end{equation} 
\end{minipage}
\hfill
\begin{minipage}[t]{0.25\textwidth}
\begin{equation}
 \nabla_j \mathcal{I}_{z} = \frac{\partial \mathcal{I}_{z}}{\partial j}
\end{equation}
\end{minipage}

Here, $\nabla_i \mathcal{I}_{z}$ and $\nabla_j \mathcal{I}_{z}$ are the partial derivatives of $\mathcal{I}_z$ with respect to $i$ and $j$. Both derivatives describe the local directional changes of intensity values in their respective horizontal or vertical direction. 

\paragraph{Magnitude image} Based on the gradient vector image the strength of the local directional changes can be quantified. The resulting image is known as the magnitude image ($\mathcal{I}^{\text{M}}_z$) and can be defined as follows:
\begin{equation}\label{eq:magnitudeimage}
	\mathcal{I}^{\text{M}}_z = |\mathcal{I}^{\nabla}_z| = \sqrt{(\nabla_i \mathcal{I}_{z})^2 + (\nabla_j \mathcal{I}_{z})^2}
\end{equation}

\paragraph{Orientation image} The orientation image ($\mathcal{I}^{\text{G}}_z$) can also be derived from the gradient vector image and can be interpreted as a complement to the magnitude image. 
While the magnitude image describes the strength of each local change, the orientation image describes the direction of each local change.

\begin{equation}\label{eq:orientationimage}
	\mathcal{I}^{\text{G}}_z = \frac{180}{\pi}\ tan^{-1}\left[\frac{\nabla_i \mathcal{I}_{z}}{\nabla_j \mathcal{I}_{z}}\right] + 180
\end{equation}

\paragraph{Vectorization} The vectorization of an image $\mathcal{I}^{\text{V}}_z: \mathcal{I}_z^{(H \times W)} \rightarrow \mathcal{I}_z^{(|\rho|)}$ describes the process to transform the two-dimensional image $\mathcal{I}_z^{(H \times W)}$ into a one-dimensional vector $\mathcal{I}_z^{(|\rho|}$, which contains only spectral positions. A common method, which is also applied in this work is the ``stacking of pixel rows'':
\begin{equation}\label{eq:spatial:sorc:1d}
	\mathcal{I}^{\text{V}}_z = \big[\mathcal{I}_{0,0,z}, \ldots, \mathcal{I}_{0, j, z}, \ldots , \mathcal{I}_{i, j ,z}, \ldots \big]
\end{equation}

\paragraph{Image histogram} Histograms of intensity values are another popular tool to analyse images. The one-dimensional histogram of a two-dimensional image $\mathcal{I}_z^{(H \times W)}$, over all spectral positions, is referred to as $P(\mathcal{I}_z)$.

\subsection{Image Scale-Space Representations}
Image scale-space representation are used to analyse structures and details at different granularity in the image \cite{witkin1987scale,lindeberg1994scale,lindeberg2007scale}. Lower scale features emphasize finer structures (like edges, corners and hot spots), while higher scales should emphasize coarser structures (like homogeneous regions). As biological tissue is organized and structured on several scales, scale space representations appear well motivated. 

The scale-space representation is a family of images generated by repeated convolution with a two-dimensional Gaussian kernel ($g$), where each repetition defines a single scale level ($t$). Formally, a scale-space representation can be defined as:
\begin{align}\label{eq:scalespace}
\begin{split}
    g(i,j;t) &= \frac{1}{2 \pi t}e^{-\frac{{i}^2+{j}^2}{2t}} \\
    L(i,j;t) &= g(i,j;t) * \mathcal{I}_{z}, \text{\ with} * \text{denoting a convolution.}
\end{split}
\end{align}
The ``;'' in $g$ and $L$ denotes a convolution that is performed over all pixels $h$ and $w$, while $t$ is a constant for the predefined scale level. Two special cases for $t$ are $t=\sigma^2$, which corresponds to the convolution with a standard Gaussian filter and $t=0$, which corresponds to the identity transformation $L(i,j;0) = \mathcal{I}_{z}$. In this manuscript Equation~\ref{eq:scalespace} is adjusted by setting $t=\sigma^2$. The step size variable $s \in \mathbb{N}^0$ is introduced and the definition of $g(i,j;t)$ is changed to $g(i,j;s\sigma^2)$ and $L(i,j;t)$ changed to $L(i,j;s\sigma^2)$. $L(i,j;s\sigma^2)$ denotes a $s$-fold convolution of $g(i,j;\sigma^2) * \mathcal{I}_{z}$.

To illustrate the influence of scale-space representations, three sets of different step sizes are used in this manuscript: $(a)\; s \in \{0\}, (b)\; s \in \{0,1,2\}$ and $(c)\; s \in \{0,2,4\}$

The main idea behind considering these three sets is, that the original scale should be more suitable for fine detailed structures like edges, corners and hot spots, while higher scales should be more suitable for coarser structures like homogeneous regions. The omission of a scale-space representation, i.e. \ set $(a.)$ above, is used as some kind of control to examine whether there is any value using scale-space representations. Utilizing sets with two different step sizes, i.e. (b.) and (c.), will examine whether there is a potential value using different scale-space levels.

\subsection{Pattern Regularity of Mass Channel Images}\label{sec:sorc:regularity}
\begin{figure}[htb!]
    \centering
    \includegraphics[width=0.96\textwidth]{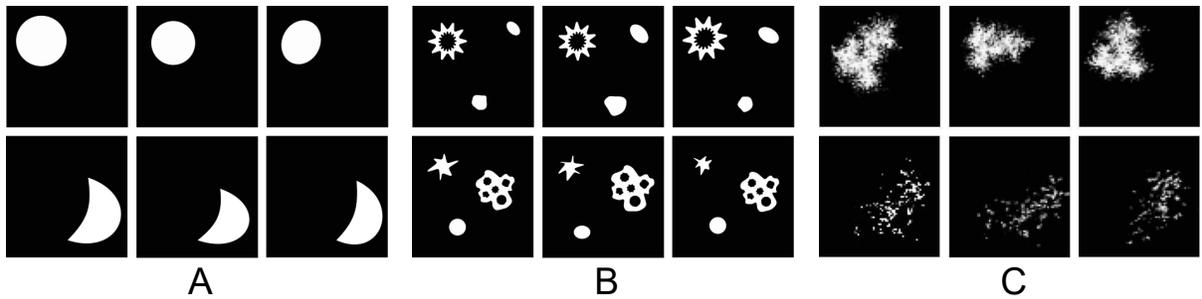}
    \caption{\textbf{Pattern regularity example} -- Three artificial image set of abstract shapes to exemplify high (A), medium (B) and low (C) regularity. The three sets consist of six images each. Images of the upper and lower row for each example are considered to build a cluster.
    (A) Artificial images with high regularity. All signal distribution pattern are areal, crisp, well defined and the pattern of the upper and lower row are well disjunct. (B) Artificial images with medium regularity. The patterns are finer and less areal, but still well defined. The pattern of the upper and lower row partially overlap, but due to their difference in shape they are still well disjunct. (C) Artificial images with low regularity. The patterns are noisy, filigree and not well defined. The pattern of the upper and lower row are still disjunct, but not as clearly as in (A).}
    \label{fig:image-complexity-examples}
\end{figure}

There is no terminology for describing morphology and structure in a biological sample beyond single chosen contexts (like histopathology classification schemes). And there also is no generally accepted terminology for the description of intensity patterns in mass channel images. In some discussions, morphologies and structures are described with one-dimensional quality scales such as \emph{simple, regular, partly regular, irregular} or \emph{complex} \cite{rezaeilouyeh2016microscopic,thiran1996morphological,yuan2006image,guo2016pathology,xing2016robust}.
Following this observation we also apply such a scale to describe and rank the structure in biological samples. The scale ranges from the term \emph{regular} to the other end denoted with \emph{irregular}.
To describe the term \emph{regularity} we propose and use the following expression: \emph{The degree of regularity in a pattern is related to the effort taken to describe the pattern, i.e. with the number of points, lines or geometrical shapes necessary to describe it.}
To give an intuition for the assessment of different levels of regularity using this expression, Figure~\ref{fig:image-complexity-examples} illustrates examples of three different levels of regularity using abstract shapes.

We use the proposed concept to describe the irregularity in Section~\ref{sec:sorc:datasets} below to categorize the samples and related data sets used in our experiments.

\subsection{Data sets}\label{sec:sorc:datasets}
Three data sets $\{\mathcal{I}^\text{B}, \mathcal{I}^\text{U}, \mathcal{I}^\text{S}\}$ from different kinds of samples (a detailed description of the data sets is given below) were chosen to represent different degrees of regularity in morphological structures, with $\mathcal{I}^\text{B}$ showing the most regular structures and $\mathcal{I}^\text{S}$ showing the least regular structures.
The three data sets were pre-processed with an in-house software, including signal alignment, normalization, matrix signal removal and peak picking. 

\subsubsection*{Barley Seed}
A barley seed cross section with a thickness of 14~\textmu m was measured with a MALDI-ToF/ToF ultrafleXtreme from Bruker Daltonics in positive ion mode. The laser diameter (pixel width) was set to 50~\textmu m, with a laser energy of 44\% to 47\% and 300 laser shots per spot. DHB was used as matrix. The ion images have a dimension of $111 \times 52$ ($H \times W$) pixels, with 3422 spectral pixels. Peak picking resulted in a set of 101 mass channel images, referred to as $\mathcal{I}^\text{B}$. The barley seed consists of a small number of compartments, which are disjunct and well defined.

\subsubsection*{Mouse Urinary Bladder}
The mouse urinary bladder data set is an example file from the \url{ms-imaging.org} website. The tissue section has a thickness of 20~\textmu m and was measured with an AP-SMALDI LTQ Orbitrap Discovery from Thermo Scientific, in positive ion mode, with a laser diameter of 10~\textmu m (pixel width). DHB was used as matrix and applied with a pneumatic sprayer. More details about the preparation of this sample can be found in \cite{rompp2010histology}. The ion images have a dimension of $260 \times 134$ ($H \times W$) pixels, with 34840 spectral pixels. Peak picking resulted in a set of 150 mass channel images, referred to as $\mathcal{I}^\text{U}$. The morphological structures in the mouse urinary bladder data are also well defined and many of the intensity patterns are disjunct. However, some of the morphological structures are filigree, which places this sample further towards the middle of the regularity spectrum . 

\subsubsection*{Human PXE Skin}
A female human pxe diseased skin section was donated after surgery and measured with a MALDI-ToF/ToF ultrafleXtreme from Bruker Daltonics in
positive mode. The laser diameter (pixel width) was set to 20~\textmu m and DHB was used as matrix. The ion images have a dimension of $395 \times 277$ ($H \times W$) pixels, with 75042 spectral pixels. Peak picking resulted in a set of 51 mass channel images, referred to as $\mathcal{I}^\text{S}$. The human Pseudoxanthoma elasticum (pxe) skin sample has a more irregular morphology. The molecular patterns, i.e.\ intensity distributions, are blurred and they overlap. In addition, some of them are wide-spread, while others are quite small and densely structured.

\subsection{SoRC Workflow}
The proposed SoRC (\textbf{s}imilarity \textbf{o}f \textbf{r}anked \textbf{c}luster indices) methodology consists of three parts: I. Pre-prorcessing (i.e.\ transformation and enhancement of data), II. Analysis (i.e.\ mass channel similarity computation (II.1) and clustering (II.2)) and III. Evaluation (i.e.\ computation of the final SoRC scores).

\begin{figure}[htb!]
	\includegraphics[width=0.9\textwidth]{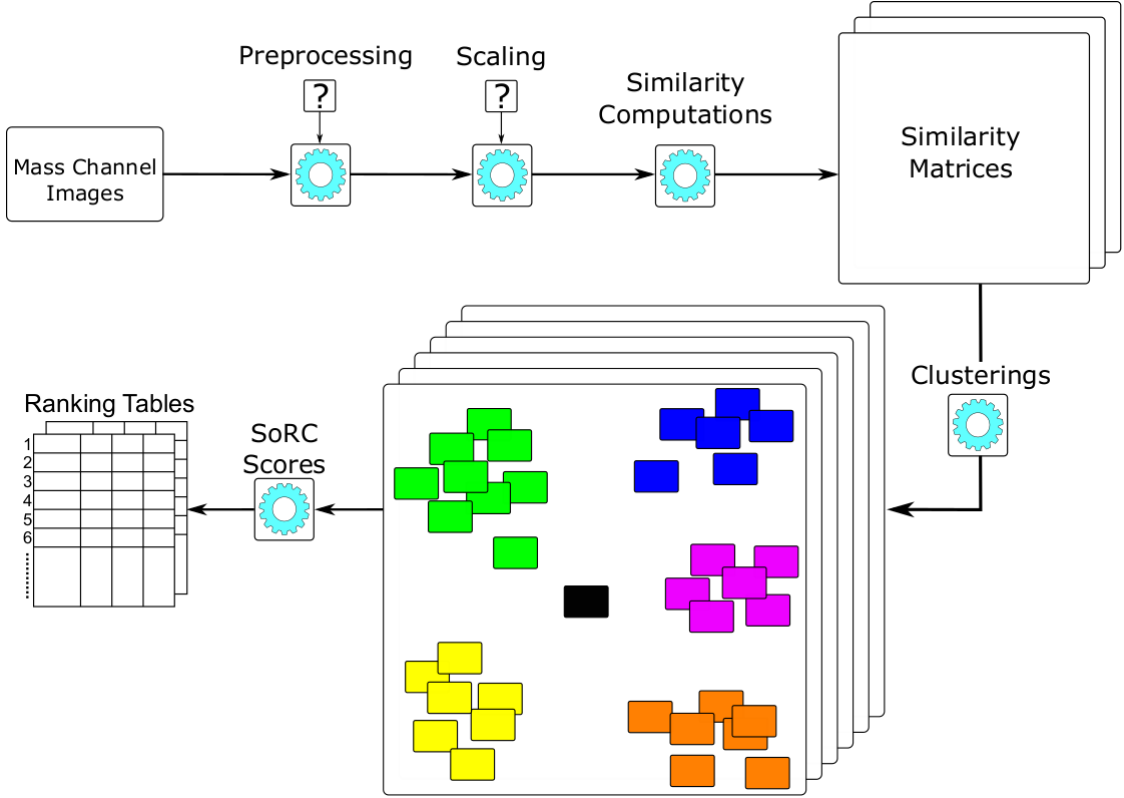}
	\caption{\textbf{SoRC workflow} -- 
	Outline of the SoRC workflow for the task of mass channel image clustering. The workflow evaluates multiple pipeline setups to propose the most suitable ones.
	To cluster mass channel images the computation of a similarity matrix and the application of a clustering procedure are required. The application of pre-processing and computation of scale-space representations is optional.
	A SoRC score is computed for each pipeline setup to suggest suitable combinations of algorithms for a given data set. The results are visualized as bar-chart-tables, i.e.\ ranking tables, combined with a bar-chart visualization.}
	\label{fig:sorc:sorc-workflow-outline}
\end{figure}

\noindent
In part I and II, different parameter settings of a modular pipeline to pre-process and cluster mass channel images are executed. As motivated above, the aim is to find clusters of mass channel images that show similar lateral molecular localization, i.e. similar signal distribution patterns. The implementation as used in this study, with clustering of mass channel images as analysis objective, is presented in Figure~\ref{fig:sorc:sorc-workflow-outline}.
Part I is subdivided into two procedures: ($P_1$) thresholding and smoothing and ($P_2$) computation of scale-space representations. The latter is carried out in two variants, referred to as $P_{2a}$ and $P_{2b}$ (see details below). The two kinds of pre-processings are tested and combined in six different setups in our study: \emph{no pre-processing}, $P_1$, $P_{2a}$, $P_{2b}$, $P_1 + P_{2a}$ and $P_1 + P_{2b}$.

The second part II includes the algorithmic process to achieve to required analytical output, in this case the computation of the similarities between pairs of mass channel images and subsequent clustering. This step uses three different clustering algorithms in combination with 13 different similarity functions (see details below). Together with the six different setups from part I sets this results in $6 \times 13 \times 3 = 234$ different clustering results. However, since the scale-space representations are not utilized by every similarity function, the actual number of different results is lowered to $138$, which is still too much to evaluate manually at the end.

The third part III consists of the computation and visualization of SoRC scores for each result computed in II. The visualization is implemented as a bar-chart integrated within a table. This tabular visualization will be referred to as bar-chart-table.

As stated in the introduction, the intention of this work is to propose a new methodology for evaluating computational pipelines and to promote one pipeline to optimal in some regard. In the following, the single computational steps of part I to III are described in more detail:  

\subsubsection{Part I: Pre-processing}
The following two pre-processing procedures are applied: 

\begin{itemize}
\item[$P_1$:] Thresholding and smoothing.
    \begin{itemize}
        \item [i.] Each mass channel images is zero-padded with a padding width of thirteen pixels.
	    \item [ii.] The intensity values are thresholded by Otsu's method \cite{otsu1979threshold}.
	    \item [iii.] Gaussian smoothing with a kernel size $\sigma = 0.8$ is applied.
	\end{itemize}
\item[$P_2$:] scale-space computation.
    \begin{itemize}
        \item [i.] Each mass channel images is zero-padded with a padding width of thirteen pixels.
    	\item [ii.] Computation of two scale-space representations according to Equation~\ref{eq:scalespace} using $s \in \{0,1,2\}$ ($P_{2a}$) and $s \in \{0,2,4\}$ ($P_{2b}$).
    \end{itemize}
\end{itemize}

\subsubsection{Part II.1: Analysis -- Similarity Functions}
We investigate thirteen different functions that cover multiple different approaches. The terms in brackets are used in the results section to relate to the individual function applied. The descriptions in this section are intentionally short and informal to provide an intuition about the concepts of the various functions. Detailed definitions can be found in Supplementary~S1.

\paragraph{Multiscale Pearson Correlation Coefficient (Pearson):}
The \emph{Pearson Correlation Coefficient} is one of the most commonly used similarity functions. Since it is defined for one-dimensional vectors or arrays we apply the transformation $\mathcal{I}_z \rightarrow \mathcal{I}_z^{\text{V}}$ according to Equation~\ref{eq:spatial:sorc:1d}. If a scale-space representation is applied to the mass channel images before, the \emph{Pearson Correlation Coefficient} is computed for each scale level individually.

\paragraph{Multiscale Cosine Similarity (Cosine):}
The \emph{Cosine Similarity} is widely applied and also originally defined for one-dimensional vectors, so we again apply the transformation $\mathcal{I}_z \rightarrow \mathcal{I}^{\text{V}}_z$. The application to scale-space representation is the same as described for the \emph{Pearson Correlation Coefficient} above.

\paragraph{Multiscale Angular Similarity (Angular):}
The \emph{Angular Similarity} is similar to the \emph{Cosine Similarity}, but more sensitive to differences between small angles. The scale-space representation handling is the same as for the \emph{Pearson Correlation Coefficient}.

\paragraph{Multiscale Structural Similarity Index (MSSIM):}
The \emph{Structural Similarity Index} (SSIM) uses local statistics to compare luminance, contrast and structure within the image and combines the result to a similarity index. In its original form, the function is defined for a pair of sliding windows to
compute a similarity map for a pair of images. Due to the sliding window approach the local neighborhood (Moore neighborhood) of each pixel is incorporated. An arithmetic mean pooling operation is applied to compute the single similarity value. 

\paragraph{Multiscale Multifeature Similarity Index (MMFS):}
Similar to the SSIM approach, the \emph{Multifeature Similarity Index} (MFS) is defined between two sliding windows. It follows the basic formula of the SSIM. However, it does not only consider the pixel intensity values, but also the luminance, contrast and structure features of the orientation images and the magnitude images (see above). So for each pair of mass channel images three similarity maps are computed and fused into one similarity . To combine them into a single similarity map, a pooling function is applied at each pixels position. In order to consider only the most dominant features, maximization is the pooling function of choice. However, the pooling function can be exchanged if desired. Again, the final similarity map has to be pooled to compute a single similarity value. Like in the SSIM, the arithmetic mean is selected.

\paragraph{Shared Pixel Information (Shared Pixel):}
The \emph{Shared Pixel Information} was implemented as a fast pixel-to-pixel comparison function that focuses only on the actual intensity values.

\paragraph{Contingency Similarity Index (Contingency):}
The \emph{Contingency Similarity Index} was implemented to investigate the suitability of multi-thresholding.

\paragraph{Hypergeometric Similarity Measure (Hypergeometric):}
Since a previous study reported very good results of the \emph{Hypergeometric Similarity Measure}  \cite{kaddi2011hypergeometric} we tested this measure as well, however in a slightly modified version that achieved better results on our data compared to the originally proposed formula.

\paragraph{Local Standard Deviation based Image Quality Index (Local Std)}
An adjusted version of the \emph{Local Standard Deviation based Image Quality Index} \cite{gore2015full} is included to investigate the effect of an intensity deviation based approach.

\paragraph{Intensity-Magnitude-Angle Similarity (IMA Sim):}
The \emph{Intensity-Magnitude-Angle Similarity} is very similar to the \emph{Multifeature Similarity Index}, but it is based a pixel-to-pixel comparison instead of a sliding window approach. Therefore, the function is faster to compute, but no local neighborhood information is incorporated (except for the gradient and magnitude information).

\paragraph{Gradient Information (Grad Info)}
The \emph{Gradient Information} function is included as a function that considers just the gradient information, i.e.\ the gradient direction and gradient magnitude, so the intensity of the signal is ignored.

\paragraph{Intensity Histogram Similarity (Histogram)}
The \emph{Intensity Histogram Similarity} compares intensity value histograms ($P(\mathcal{I}_z)$) of mass channel images using the Hellinger distance \cite{hellinger1909neue} or another function. In this work we apply the Hellinger distance \cite{hellinger1909neue}, as it is directly related to the Euclidean norm, which supports an intuitive understanding of the comparison approach.

\paragraph{Mutual Information (Mutual Info)}
\emph{Mutual Information}
is a commonly known information theoretical measure that quantifies the ``amount of information''. It is included since information theoretical approaches are regularly used in many different areas of signal analysis \cite{pluim2003mutual}.

\subsubsection{Part II.2: Analysis -- Clustering Methods}
The following three clustering algorithms are applied in this manuscript. All three methods are briefly described with more details in the supplementary (S2). 
\noindent
\textbf{Agglomerative hierarchical clustering} \cite{sokal1958statistical} is very prominent in bioinformatics and applied in many contexts. Data points are iteratively joint into clusters according a selected similarity criterion until all data points are joint into one cluster. In this work we apply the \emph{agglomerative average linkage} version \cite{sokal1958statistical}.  

\noindent
\textbf{Affinity propagation} \cite{frey2007clustering} determines a set of data points that best suited to represent the entire data in an iterative way, by exchanging messages between points evaluating the suitability of a point to be representative for others. 

\noindent
\textbf{Community detection} \cite{wullems2019detection} clusters data points based on the evaluation of a graph representation. The graph is based on an adjacency matrix computed from a similarity matrix computed from the points' coordinates. 

Besides these three clustering algorithms, the publicly available SoRC workflow implementation currently supports $k$-medoids (partitioning around medoids (PAM) version \cite{novikov2019pyclustering} and expectation-maximization (EM) version \cite{de2003c}), DBSCAN \cite{ester1996density}, OPTICS \cite{ankerst1999optics} and Spectral clustering \cite{shi2000normalized}. But the workflow can easily be extended to include any type of clustering procedure, as long as the labels are adjusted to start with index one and noise labels, if they exist, are assigned to unique numbers.

\subsubsection{Part III: Evaluation -- SoRC Score}
The SoRC score is designed to quantify the quality of a clustering result with a single number. The computation of a single score allows an easy visualization, ranking and comparison for a multitude of chosen methods and modifications. This way it increases the efficiency in the process of designing and improving a software pipeline for grouping mass channel images into clusters.

A clustering is considered to be of a higher quality if the result clusters are dense, well separated and represent the most important structural features of the data. The SoRC score uses two different cluster indices, which are not strictly correlated, to express these properties in numbers: The Silhouette Coefficient score (SCS) \cite{rousseeuw1987silhouettes} and the Calinski-Harabasz index (CHI) \cite{calinski1974dendrite}

\noindent The Silhouette Coefficient SCS$\in [-1,1]$ is based on a comparison of a data point's position in relation to its assigned cluster (\emph{cohesion}) and to the next (or second best) nearest other cluster (\emph{separation}). The arithmetic mean over all data points is taken as final SCS value. A higher number of unambiguous data point - cluster assignments results in a higher number of SCS values. So $SCS=1$ indicates a coherent and well separated (i.e. non-ambiguous) cluster assignment. Lower values indicate lesser separation and lesser coherence. 

The Calinski-Harabasz index (CHI)$\in [0, \infty]$ is defined as the ratio between the within-cluster dispersion and the between-cluster dispersion. The index increases with decreasing intra-cluster variance and increasing inter-cluster distance.

The main difference between these two cluster coefficients (or indices) is that the SCS is computed on the actual similarity values, while the CHI is computed directly on the data points within the clusters. So an integration of both values into one SoRC score enables the integration of both aspects, the composition of the similarity values and the composition of the images (i.e. data points) within each cluster.

Due to their different output ranges (see above) we do consider the rank and not the actual value of the values. So for both measures, the final values are ranked and the rank position values are normalized into $[0,1]$. Consequently, higher ranks represent better clusters. Ties are solved by mean ranks, e.g. the values $[1, 0.7, 0.7, 0.7, 0.4]$ rank to $[5,3,3,3,1]$. The actual SoRC score is the arithmetic mean of the normalized ranks of both cluster indices.

\section{Results}
\begin{figure}[htb!]
	\centering
	{\large Mouse Urinary Bladder ($\mathcal{I}^{U}$) with Agglomerative Hierarchical Clustering}
	
	\vspace{\baselineskip}
	
	\includegraphics[width=0.9\textwidth]{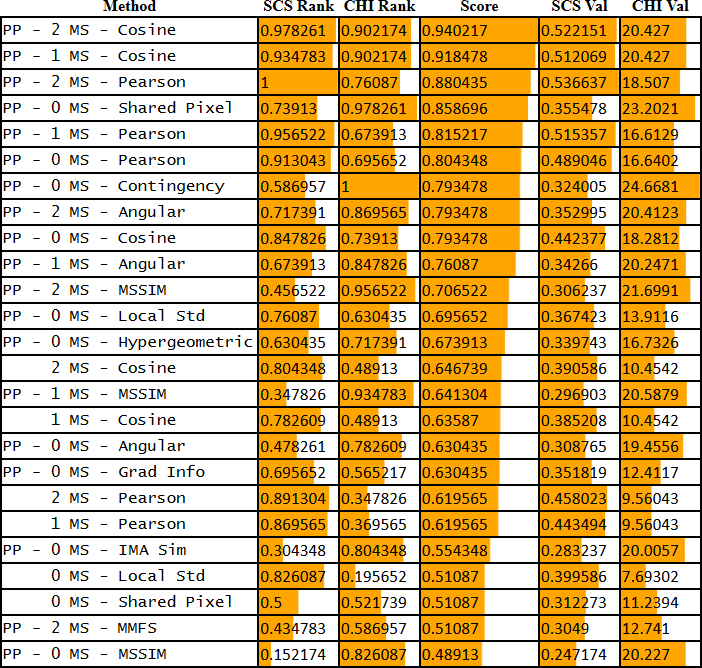}
	\caption{\textbf{SoRC score bar-chart-table example} -- The bar-chart-table visualization shows the $25$ highest SoRC scores (out of a total of $46$) for the mouse urinary bladder ($\mathcal{I}^{U}$) in combination with agglomerative hierarchical clustering. The \textbf{Method} column contains the applied combination of pre-processing, scale-space representation and similarity function. \texttt{PP} indicates pre-processing and \texttt{0MS}, \texttt{1MS} and \texttt{2MS} indicate the utilization of scale-spaces with $s=0$, $s=1$ and $s=2$, respectively. The columns \textbf{SCS Rank} and \textbf{CHI Rank} contain the normalized ranks. The ranks are computed on the basis of the absolute values of the respective cluster index, which are shown in the columns \textbf{SCS Val} and \textbf{CHI Val}. The whole bar-chart-table is sorted according to the SoRC scores, which are shown in the \textbf{Score} column.}
	\label{fig:sorc:sorc-scores-example-bladder}
\end{figure}

Before all results are compared with each other, we present an example for the entire evaluation procedure using the agglomerative hierarchical clustering results for the mouse urinary bladder data set ($\mathcal{I}^{U}$). Figure~\ref{fig:sorc:sorc-scores-example-bladder} shows the bar-chart-table visualization for the different combinations in part I and II (pre-processing and  similarity function) but with a fixed clustering algorithm (hierarchical) with the $25$ highest SoRC scores (out of a total of $46$). It can be seen that in these $25$ examples, functions with pre-processing always score better than their counterparts without pre-processing. For those similarity functions where the integration of a scale-space is available, the larger step size scores better than the smaller step size, which scores better than no scale-space.
The evaluation of the individual columns for the SCS and the CHI shows that the indices ``disagree'' for some combinations, i.e.\ one index rates the clustering better than the other. This demonstrates the importance of using two not strictly correlated cluster indices that assess different characteristics of a clustering. In general, the SoRC score is higher if both indices rate the clustering similarly. The original values, presented in the last two columns, can be used to either examine the quality difference between two pipeline setups in more detail or to compare the bar-chart-tables for different clustering procedures.

\begin{figure}[htb!]
	\begin{minipage}[t]{1\textwidth}
		\begin{minipage}[t]{0.03\textwidth}
		\
	    \end{minipage}
		\begin{minipage}[t]{0.305\textwidth}
			\begin{center}
				Affinity Propagation
			\end{center}
	    \end{minipage}
		\begin{minipage}[t]{0.305\textwidth}
			\begin{center}
				Hierarchical Clustering
			\end{center}
	    \end{minipage}
	    \begin{minipage}[t]{0.305\textwidth}
			\begin{center}
				Community Detection
			\end{center}
	    \end{minipage}
    \end{minipage}
    
    \vspace{\baselineskip}
    
    \begin{minipage}[t]{1\textwidth}
		\begin{minipage}[b]{0.03\textwidth}
			$\mathcal{I}^{B}$ \\
			\ \\
	    \end{minipage}
		\begin{minipage}[t]{0.305\textwidth}
			\includegraphics[width=\textwidth]{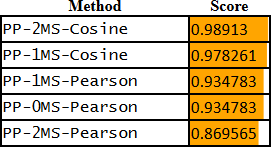}
	    \end{minipage}
		\begin{minipage}[t]{0.305\textwidth}
			\includegraphics[width=\textwidth]{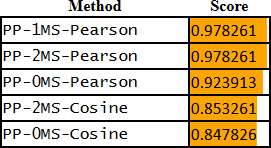}
	    \end{minipage}
	    \begin{minipage}[t]{0.305\textwidth}
			\includegraphics[width=\textwidth]{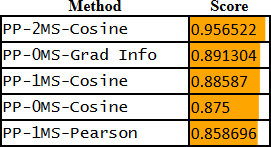}
	    \end{minipage}
    \end{minipage}
    
    \vspace{\baselineskip}
    
    \begin{minipage}[t]{1\textwidth}
		\begin{minipage}[b]{0.03\textwidth}
			$\mathcal{I}^{U}$ \\
			\ \\
	    \end{minipage}
		\begin{minipage}[t]{0.305\textwidth}
			\includegraphics[width=\textwidth]{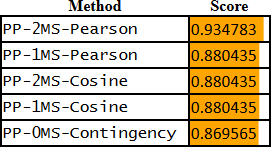}
	    \end{minipage}
		\begin{minipage}[t]{0.305\textwidth}
			\includegraphics[width=\textwidth]{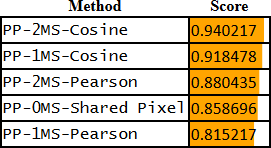}
	    \end{minipage}
	    \begin{minipage}[t]{0.305\textwidth}
			\includegraphics[width=\textwidth]{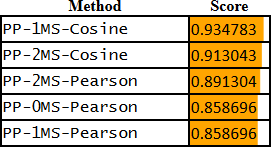}
	    \end{minipage}
    \end{minipage}
    
    \vspace{\baselineskip}
    
    \begin{minipage}[t]{1\textwidth}
		\begin{minipage}[b]{0.03\textwidth}
			$\mathcal{I}^{S}$ \\
			\ \\
	    \end{minipage}
		\begin{minipage}[t]{0.305\textwidth}
			\includegraphics[width=\textwidth]{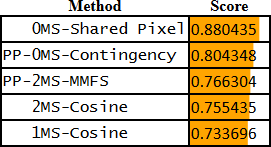}
	    \end{minipage}
		\begin{minipage}[t]{0.305\textwidth}
			\includegraphics[width=\textwidth]{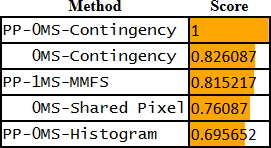}
	    \end{minipage}
	    \begin{minipage}[t]{0.305\textwidth}
			\includegraphics[width=\textwidth]{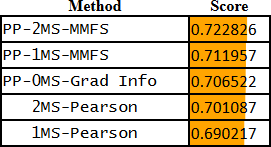}
	    \end{minipage}
    \end{minipage}
	\caption{\textbf{SoRC score comparison} -- Comparison of the top five computed SoRC scores for each combination of data set and cluster method. \texttt{PP} indicates pre-processing. \texttt{0MS}, \texttt{1MS} and \texttt{2MS} indicate the utilization of scale-spaces with $s=0$, $s=1$ and $s=2$, respectively. A higher score at the same position does not necessarily imply a better result. It rather indicates that there are less numerical ties and less disagreements between SHS and CHI.}
	\label{fig:sorc:sorc-scores-comparison}
\end{figure}

To finally compare the different clustering methods in the analysis, multiple bar-chart-tables can be compared, as presented in Figure~\ref{fig:sorc:sorc-scores-comparison}. HTML files with bar-chart-tables for each individual combination of data set and clustering method are provided as part of the Additional Files. To provide a clear comparison, the columns of the bar-chart-tables are reduced to the SoRC scores and the rows are reduced to the five highest scoring context combinations. The evaluation of the comparison yields five major observations:
\begin{enumerate}
	\item The \emph{Cosine Similarity} and the \emph{Pearson Correlation Coefficient} are among the best scoring similarity functions for most combinations of data set and clustering method. The only exception is the human skin data set ($\mathcal{I}^{S}$). 
	\item For $\mathcal{I}^{B}$ and $\mathcal{I}^{U}$, most similarity functions achieve better results in combination with pre-processing. 
	\item The results from our study indicate, that similarity measures based clustering can achieve better results when they include scale-space representations. But there is no clear result whether the smaller or the larger step size performs better. 
	\item The five highest SoRC scores for $\mathcal{I}^{B}$ and $\mathcal{I}^{U}$ are higher than for $\mathcal{I}^{S}$ in almost all cases. This shows that the selected cluster indices quantify the clustering quality of the irregular sample differently more often than it is the case for samples with higher regularity.
	\item Global image features, discarding the local features achieve inferior results which indicates that discarding any positional relationship between the pixels is likely to have a negative impact on the clustering result.
	\item The last and probably most important observation is that no pipeline setup always leads to the best score.
\end{enumerate}

Table~\ref{tab:time-requirements} shows a summary of the time requirements needed to evaluate each data set with the SoRC workflow. Each pre-processing setup was executed in parallel. It can be seen that with increasing number and size of images the time requirements increase notably. This is even more important if the different pre-processing setups are not executed in parallel.

\begin{table}
	\caption{
	\textbf{Time requirements} -- Time requirements for the whole SoRC workflow. Each pre-processing setup was executed in parallel on a cloud machine with 500 GB memory and 28 CPUs. \textbf{Total} illustrates the time requirement if no parallel execution is used.}
	\begin{tabular}{rrrr}\toprule
		 Pre-processing Setup & \multicolumn{1}{c}{Barley} & \multicolumn{1}{c}{Bladder} & \multicolumn{1}{c}{Skin} \\ \cmidrule(rl){1-1} \cmidrule(rl){2-2} \cmidrule(rl){3-3} \cmidrule(rl){4-4}
		0 MS & 1:08:58 & 4:05:14 & 1:00:43 \\ 
		1 MS & 1:20:42 & 4:36:36 & 1:08:46 \\ 
		2 MS & 1:21:21 & 4:36:52 & 1:10:47 \\ 
		PP -- 0 MS & 1:08:04 & 3:17:23 & 0:42:36 \\ 
		PP -- 1 MS & 1:19:03 & 3:57:17 & 0:50:53 \\ 
		PP -- 2 MS & 1:19:59 & 3:57:14 & 0:50:47  \\ \cmidrule(rl){1-4}
		Total & 7:38:07 & 24:30:36 & 5:44:32  \\ \bottomrule
	\end{tabular}
	\label{tab:time-requirements}
\end{table}

\section{Discussion}
Using our workflow and the SoRC visualizations we were able to observe, that pre-processing (de-noising, multi scale representation) seems to have a positive impact on the final clustering results, independent from the the clustering method or the similarity function. Please note, that we do not want to generalize this observation to data outside this study. On the one hand, de-noising pre-processing can have a positive influence on commonly known MSI problems, such as pixel-to-pixel variation and hotspots. On the other hand, it can have a negative influence on molecular distributions that are either small or finely structured. The same applies to the integration of different scale-space levels. The idea to combine multiple levels to consider fine grained structures at lower scales and coarser structures at higher scales seems to work out well in most cases. An integration of this approach as presented in this manuscript, i.e.\ by combining the results of several scale levels with a mean function, can theoretically be used in combination with any similarity function.

Since we did not find such a comparative analysis for similarity functions and different degrees of sample complexity in the literature one motivation for SoRC was the comparative analysis of different similarity functions. The results confirmed that two of the most commonly used functions (\emph{Pearson Correlation Coefficient} and \emph{Cosine Similarity}) are good choices, which is especially true for samples with a more regular structure. Interestingly, this result is in line with the recommendation given in the work of Ovchinnikova et al. \cite{ovchinnikova2020colocml}. This previous comparison also recommends to use pre-processing in comparison with the \emph{Cosine Similarity}. To be precise, they recommend to use a median signal thresholding and a sliding window median filter of size three, which also supports our statement about the importance of the pipeline setup.

However, it was also revealed that the choice of a suitable similarity measure becomes more difficult with increasing irregularity of the sample. This is supported by the observation that alternative image features, such as gradients or magnitudes, seem to gain in importance with increasing irregularity. The same seems to apply to the preservation of local neighborhoods.

All results taken together it becomes clear, that evaluation workflows with numerical quality estimations provide a great support to increase the quality of analysis pipelines. 

\section{Conclusion}
In this manuscript, the suitability of different similarity functions to evaluate the co-localization of mass channel images was investigated. The influence of pre-processing and image scales was considered and the interaction of data source, analysis pipeline and analysis objective was addressed. Additionally, the SoRC evaluation workflow was presented. The workflow proposes a method to estimate the quality of an analysis pipeline by using a single score, the SoRC score. Based on this score, the user is provided with suggestions for combinations of methods. The use of the SoRC workflow allows to evaluate multiple pipeline setups fully automatically. This saves a time-consuming execution and evaluation of each pipeline setup and its parameters individually (see Table~\ref{tab:time-requirements}). In addition, the otherwise necessary tedious pairwise comparison of the results is avoided. Consequently, streamlining the comparison of pipeline setups through the SoRC workflow significantly improves the efficiency of the evaluation.
SoRC in its present form is only suitable to evaluate clustering as analysis objective and its execution requires some programming knowledge. However, the underlying methodology can be extended to any analysis objective, as long as one can define at least one numerical value to estimate the quality of a result.
However, if time or computational resources are limited we advice to use either the  \emph{Pearson Correlation Coefficient} or the \emph{Cosine Similarity} in combination with basic pre-processing techniques and the integration of a small scale-space representation. Chances are high that a combination of these methods and functions provides good results for regular structured samples and at least acceptable results for irregular structured samples.

\section*{Availability}
Source code is available on github.\\
\textbf{Project name:} SoRC \\
\textbf{Project home page:} \href{https://github.com/Kawue/sorc}{https://github.com/Kawue/sorc} \\
\textbf{Operating system(s):} Platform independent \\
\textbf{Programming language:} Python \\
\textbf{Other requirements:} Python 3.5 or higher \\
\textbf{License:} GNU GPLv3 \\

\section*{Author's contributions}
    Concept and design of study and software: KW, TWN. Developed and implemented software: KW. Analyzed data: KW. Evaluation and interpretation of results: KW, TWN. Writing: KW, TWN. All authors read and approved the final manuscript.

\section*{Funding}
  KW was funded by the International DFG Research Training Group GRK 1906. The funding body played no role regarding the design of the study, data collection, analysis or interpretation or the writing of the manuscript.
  
\section*{Competing interests}
  The authors declare that they have no competing interests.
  
\section*{Ethics approval and consent to participate}
The collection and use of the human skin tissue was approved by the ethics commission of the Ruhr University of Bochum Faculty of Medicine, located in Bad Oeynhausen. All patients (donors of the human tissue) provided their informed consent in written form. We obtained the data completely anonymized.

\newpage
\renewcommand\thesubsection{S\arabic{subsection}}
\section*{Supplementary}

\subsection{Similarity Functions}

\subsubsection{Multiscale Pearson Correlation Coefficient (Pearson):}

\begin{align}
\begin{split}\label{eq:supp:sorc:functions:pearson-ms}
    &\delta_s(\mathcal{I}^\text{V}_{z;s}, \mathcal{I}^\text{V}_{z';s}) = \frac{\sigma(\mathcal{I}^\text{V}_{z;s}, \mathcal{I}^\text{V}_{z';s})}{\sigma(\mathcal{I}^\text{V}_{z;s}) \sigma(\mathcal{I}^\text{V}_{z';s})} \\
    &\mathcal{\delta} = \frac{1}{S} \sum_{0}^{S} \mathcal{\delta}_s(\mathcal{I}_{z;s},\mathcal{I}_{z';s})
\end{split}
\end{align}
where $\sigma(\cdot)$ and $\sigma(\cdot, \cdot)$ are the variance and covariance function.

\subsubsection{Multiscale Cosine Similarity (Cosine):}
\begin{align}
\begin{split}\label{eq:supp:sorc:functions:cosine-ms}
    &\delta_s(\mathcal{I}^\text{V}_{z;s},\mathcal{I}^\text{V}_{z';s}) = \frac{\mathcal{I}^\text{V}_{z;s} \cdot \mathcal{I}^\text{V}_{z';s}}{\norm{\mathcal{I}^\text{V}_{z;s}} \norm{\mathcal{I}^\text{V}_{z';s}}} \\
    &\mathcal{\delta} = \frac{1}{S} \sum_{0}^{S} \mathcal{\delta}_s(\mathcal{I}_{z;s},\mathcal{I}_{z';s})
\end{split}
\end{align}
where $\norm{\cdot}$ is the Euclidean norm.

\subsubsection{Multiscale Angular Similarity (Angular):}
\begin{align}
\begin{split}\label{eq:supp:sorc:functions:angular-ms}
    &\delta_s(\mathcal{I}^\text{V}_{z;s}, \mathcal{I}^\text{V}_{z';s}) = 1-\left(\frac{2}{\pi} cos^{-1}\left(\frac{\mathcal{I}^\text{V}_{z;s} \cdot \mathcal{I}^\text{V}_{z';s}}{\norm{\mathcal{I}^\text{V}_{z;s}} \norm{\mathcal{I}^\text{V}_{z';s}}}\right)\right) \\
    &\mathcal{\delta} = \frac{1}{S} \sum_{0}^{S} \mathcal{\delta}_s(\mathcal{I}_{z;s},\mathcal{I}_{z';s})
\end{split}
\end{align}
Equation~\ref{eq:supp:sorc:functions:angular-ms} is only valid for $\mathcal{I}^\text{V}_{z;s} \land \mathcal{I}^\text{V}_{z';s} \in \mathbb{R}_{\geq 0}$. For $\mathcal{I}^\text{V}_{z;s} \lor \mathcal{I}^\text{V}_{z';s} \in \mathbb{R}$ the factor $\frac{2}{\pi}$ must be changed to $\frac{1}{\pi}$.

\subsubsection{Multiscale Structural Similarity Index (MSSIM):}
The structural similarity function is defined between two sliding windows of size $o \times o$, with $o=13$.
\begin{align}
\begin{split}\label{eq:supp:sorc:functions:ssim-ms}
	&\kappa^{(o \times o)}(\mathcal{I}_{z;s},\mathcal{I}_{z';s}) = \frac{(2\mu(\mathcal{I}_{z;s}) \mu(\mathcal{I}_{z';s}) + c_1)(2\sigma(\mathcal{I}_{z;s}, \mathcal{I}_{z';s}) + c_2)}{(\mu^2(\mathcal{I}_{z;s}) + \mu^2(\mathcal{I}_{z';s}) + c_1)(\sigma^2(\mathcal{I}_{z;s}) + \sigma^2(\mathcal{I}_{z';s}) + c_2)}\\
    &\delta_s(\mathcal{I}_{z;s},\mathcal{I}_{z';s}) = \frac{1}{|\rho|} \sum_{i,j \in \rho} \kappa^{(o \times o)}(\mathcal{I}_{z;s},\mathcal{I}_{z';s}) \\
    &\mathcal{\delta} = \frac{1}{S} \sum_{0}^{S} \mathcal{\delta}_s(\mathcal{I}_{z;s},\mathcal{I}_{z';s})
\end{split}
\end{align}
where $\kappa^{(o \times o)}(\cdot, \cdot)$ is a sliding window function of size $o \times o$ across two images. The function returns a new image, which is called similarity map. $\mu(\cdot)$ and $\sigma(\cdot)$ calculate the arithmetic mean and the standard deviation of the sliding window, $c_1,c_2$ are two small constants to avoid division by zero and $|\rho|$ is the size of the set of all spectral positions.

\subsubsection{Multifeature Similarity Index (MMFS):}
Similar to the SSIM approach, the MFS Max is defined between two sliding windows of size $o \times o$, with $o=13$.
\begin{align}
\begin{split}\label{eq:supp:sorc:functions:mfs-max}
    &\mathcal{F}^{(o \times o)}_\mu(\mathcal{I}_{z;s},\mathcal{I}_{z';s}) = 
    \frac{2 \mu(\mathcal{I}_{z;s}) \mu(\mathcal{I}_{z';s}) +c}{{\mu(\mathcal{I}_{z;s})}^2+{\mu(\mathcal{I}_{z';s})}^2 +c} \\
    &\mathcal{F}^{(o \times o)}_\sigma(\mathcal{I}_{z;s},\mathcal{I}_{z';s}) = 
    \frac{2 \sigma(\mathcal{I}_{z;s}) \sigma(\mathcal{I}_{z';s}) +c}{{\sigma(\mathcal{I}_{z;s})}^2+{\sigma(\mathcal{I}_{z';s})}^2 +c} \\
    &\mathcal{F}^{(o \times o)}_{cov}(\mathcal{I}_{z;s},\mathcal{I}_{z';s}) = \frac{\sigma(\mathcal{I}_{z;s}, \mathcal{I}_{z';s}) + c}{\sigma(\mathcal{I}_{z;s})\sigma(\mathcal{I}_{z;s})+c} \\       
    &\Xi_I = \mathcal{F}_\mu(\mathcal{I}_{z;s}, \mathcal{I}_{z';s}) \mathcal{F}_\sigma(\mathcal{I}_{z;s}, \mathcal{I}_{z';s}) \mathcal{F}_{cov}(\mathcal{I}_{z;s}, \mathcal{I}_{z';s}) \\        
    &\Xi_O = \mathcal{F}_\mu(\mathcal{I}^\text{G}_{z;s}, \mathcal{I}^\text{G}_{z';s})\ \mathcal{F}_\sigma(\mathcal{I}^\text{G}_{z;s}, \mathcal{I}^\text{G}_{z';s})\ \mathcal{F}_{cov}(\mathcal{I}^\text{G}_{z;s}, \mathcal{I}^\text{G}_{z';s}) \\
    &\Xi_M = \mathcal{F}_\mu(\mathcal{I}^\text{M}_{z;s}, \mathcal{I}^\text{M}_{z';s})\ \mathcal{F}_\sigma(\mathcal{I}^\text{M}_{z;s}, \mathcal{I}^\text{M}_{z';s})\ \mathcal{F}_{cov}(\mathcal{I}^\text{M}_{z;s}, \mathcal{I}^\text{M}_{z';s}) \\
    &\Xi_{I'} = \text{sign}(\Xi_I) |\Xi_I|^\frac{1}{3}; \text{\hspace{0.6cm}} \Xi_{O'} = \text{sign}(\Xi_O) |\Xi_O|^\frac{1}{3}; \text{\hspace{0.6cm}} \Xi_{M'} = \text{sign}(\Xi_M) |\Xi_M|^\frac{1}{3} \\
    &\Xi_W = \max_{i,j  \in \rho}(\Xi_{I_{i,j}}, \Xi_{O_{i,j}}, \Xi_{M_{i,j}}) \\
    &\delta_s(\mathcal{I}_{z;s},\mathcal{I}_{z';s}) = \frac{1}{|\rho|} \sum_{i,j  \in \rho} \Xi_W;\\
    &\mathcal{\delta} = \frac{1}{S} \sum_{0}^{S} \mathcal{\delta}_s(\mathcal{I}_{z;s},\mathcal{I}_{z';s})
\end{split}
\end{align}
where $\mathcal{F}_\mu$, $\mathcal{F}_\sigma$ and $\mathcal{F}_{cov}$ are sliding window functions of size $o \times o$ across two images. Like $\kappa$ in Equation~\ref{eq:supp:sorc:functions:ssim-ms} these functions return similarity maps. $\mu(\cdot)$, $\sigma(\cdot)$ and $\sigma(\cdot, \cdot)$ calculate the arithmetic mean, standard deviation and covariance of the sliding window.
$\text{sign}(\cdot)$ represents the signs of its input, $|\Xi|$ represents the absolute values of $\Xi$ and $c$ is a small constants to avoid division by zero.

\subsubsection{Shared Pixel Information (Shared Pixel)}
\begin{equation}\label{eq:supp:sorc:functions:sharedresidual}
    \delta(\mathcal{I}_{z},\mathcal{I}_{z'}) = 1 - \frac{\sum_{i,j \in \rho} |\mathcal{I}_{z} - \mathcal{I}_{z'}|}{\sum_{i,j \in \rho} \mathcal{I}_{z} + \sum_{i,j \in \rho} \mathcal{I}_{z'}}
\end{equation}

\subsubsection{Contingency Similarity Index (Contingency):}
\begin{align}
\begin{split}\label{eq:supp:sorc:functions:contingency}
	&c_\text{l}(\mathcal{I}_z) = 0.2 \cdot \max{(\mathcal{I}_z)}\\
	&c_\text{u}(\mathcal{I}_z) = 0.8 \cdot \max{(\mathcal{I}_z)}\\
    &\mathcal{I}^c_{i,j,z} =
    \begin{cases}
        1,& \text{if } \mathcal{I}_{i,j,z} < c_\text{l}(\mathcal{I}_z) \land i,j \in \rho\\
        2,&  c_\text{l}(\mathcal{I}_z) < \mathcal{I}_{i,j,z} < c_\text{u}(\mathcal{I}_z) \land i,j \in \rho \\
        3,& \text{if } \mathcal{I}_{i,j,z} > c_\text{u}(\mathcal{I}_z) \land i,j \in \rho \\
        0,& \text{if } i,j \notin \rho
    \end{cases}\\
    & \mathcal{V} = \mathfrak{V}_{\{1,2,3\}}(\mathcal{I}^c_{z}, \mathcal{I}^c_{z'}) \\
	&\delta(\mathcal{I}_{z},\mathcal{I}_{z'}) = \frac{1}{\sum \mathcal{V}} \big((\mathcal{V}_{0,0} + \mathcal{V}_{1,1} + \mathcal{V}_{2,2})\\
	&\qquad - (\mathcal{V}_{0,1} + \mathcal{V}_{1,2} + \mathcal{V}_{1,0} + \mathcal{V}_{2,1}) \\
    &\qquad - (\mathcal{V}_{0,2} + \mathcal{V}_{2,0})\big) \\
\end{split}
\end{align}
where $\mathfrak{V}_{\{1,2,3\}}(\cdot, \cdot)$ calculates a $3 \times 3$ contingency table of two images for values in {\{1,2,3\}}.

\subsubsection{Hypergeometric Similarity Measure (Hypergeometric):}
\begin{align}
\begin{split}\label{eq:supp:sorc:functions:hypergeometric}
	&|\mathcal{I}_{z}| = \sum \text{binarize}(\mathcal{I}_{z})\\
    &|\mathcal{I}_{z'}| = \sum \text{binarize}(\mathcal{I}_{z'})\\
    &|\mathcal{I}_{z} \land \mathcal{I}_{z'}| = \sum \text{binarize}(\mathcal{I}_{z}) \land \text{binarize}(\mathcal{I}_{z'}) \\
	&\beta_1 = \frac{|\rho|-|\mathcal{I}_{z'}|}{|\rho|} \\
    &\alpha_1 = \frac{|\mathcal{I}_{z}|-|\mathcal{I}_{z} \land \mathcal{I}_{z'}|}{|\mathcal{I}_{z}|}-\beta_1 \\
    &\beta_2 = \frac{|\mathcal{I}_{z'}|}{|\rho|} \\
    &\alpha_2 = \frac{|\mathcal{I}_{z} \land \mathcal{I}_{z'}|}{|\mathcal{I}_{z}|}-\beta_2 \\
	& \hat{\alpha} =
    	\begin{cases}
	        {\frac{\beta_1}{\beta_1+\alpha_1}^{\beta_1+\alpha_1}  \frac{1-\beta_1}{1-\beta_1-\alpha_1}^{1-\beta_1-\alpha_1}}^2,& \text{if } 1-\beta_1-\alpha_1 > 0 \land \beta_1+\alpha_1 > 0\\
		    \infty,              & \text{otherwise}
        \end{cases} \\
    & \hat{\beta} =
		\begin{cases}
        	{\frac{\beta_2}{\beta_2+\iota_2}^{\beta_2+\alpha_2}  \frac{1-\beta_2}{1-\beta_2-\alpha_2}^{1-\beta_2-\alpha_2}}^2,& \text{if } 1-\beta_2-\alpha_2 > 0 \land \beta_2+\alpha_2 > 0\\
        	\infty,              & \text{otherwise}
    	\end{cases} \\
	&\delta' = 
    	\begin{cases}
    		1, & \text{if } \hat{\alpha} - \hat{\beta} = -\infty \\
	        -1, & \text{if }  \hat{\alpha} - \hat{\beta} = \infty \\
	        \hat{\alpha}- \hat{\beta}, & \text{otherwise}
    	\end{cases} \\
    &\delta(\mathcal{I}_{z},\mathcal{I}_{z'}) = \frac{1}{2} (\delta'(\mathcal{I}_{z}, \mathcal{I}_{z'}) + \delta'(\mathcal{I}_{z'}, \mathcal{I}_{z}))\\
\end{split}
\end{align}
where $\text{binarize}(\cdot)$ describes a binarization function.

\subsubsection{Local Standard Deviation based Image Quality Index (Local Std):}
\begin{align}
\begin{split}\label{eq:supp:sorc:functions:local-std}
    &\delta(\mathcal{I}_{z},\mathcal{I}_{z'}) = \frac{1}{|\rho|} \sum_{i,j \in \rho} \mathcal{I}_\sigma \\
    &\mathcal{I}_\sigma = \frac{2 {f_\sigma(\mathcal{I}_z)} {f_\sigma(\mathcal{I}_z')} + c}{{f_\sigma(\mathcal{I}_z)}^2 +  {f_\sigma(\mathcal{I}_z')}^2 + c}
\end{split}
\end{align}
where $f_\sigma(\mathcal{I}_z)$ is the result of the convolution of $\mathcal{I}_z$ with a standard deviation kernel of size $13$ and c is a small constant to avoid division by zero.

\subsubsection{Intensity-Magnitude-Angle Similarity (IMA Sim):}
\begin{align}
\begin{split}\label{eq:supp:sorc:functions:intmagan} 
    &\delta(\mathcal{I}_{z},\mathcal{I}_{z'}) = \frac{1}{|\rho|} \sum_{i,j \in \rho} \frac{1}{3} \delta_\text{int}(\mathcal{I}_{z},\mathcal{I}_{z'}) + \frac{1}{3} \delta_\text{mag}(\mathcal{I}_{z'},\mathcal{I}_{z'}) + \frac{1}{3} \frac{\delta_\text{ang}(\mathcal{I}_{z},\mathcal{I}_{z'})+1}{2}\\
    &\delta_\text{int}(\mathcal{I}_{z},\mathcal{I}_{z'}) = \frac{2 \mathcal{I}_{z} \mathcal{I}_{z'} + c}{{\mathcal{I}_{z}}^2 + {\mathcal{I}_{z'}}^2 + c} \\
    &\delta_\text{mag}(\mathcal{I}_{z},\mathcal{I}_{z'}) = \frac{2\ \mathcal{I}^\text{M}_{z} \mathcal{I}^\text{M}_{z'}| + c}{({\mathcal{I}^\text{M}_{z}})^2 + ({\mathcal{I}^\text{M}_{z'}})^2 + c} \\
    &\delta_\text{ang}(\mathcal{I}_{z},\mathcal{I}_{z'}) = \mathcal{I}^\nabla_{z} \cdot \mathcal{I}^\nabla_{z'}
\end{split}
\end{align}

\subsubsection{Gradient Information (Grad Info):}
\begin{align}
\begin{split}\label{eq:supp:sorc:functions:gradient-info}
    &\delta(\mathcal{I}_{z},\mathcal{I}_{z'}) = 
    \frac{1}{|\rho|} \sum_{i,j \in \rho} \delta' \min(\mathcal{I}^\text{M}_{z}, \mathcal{I}^\text{M}_{z'})\\
    &\delta' = \frac{1}{2}\left(\frac{\mathcal{I}^\nabla_{z} \cdot \mathcal{I}^\nabla_{z'}}{\mathcal{I}^\text{M}_{z} \mathcal{I}^\text{M}_{z'}} + 1 \right)
\end{split}
\end{align}

\subsubsection{Intensity Histogram Similarity (Histogram):}
\begin{align}
\begin{split}\label{eq:supp:sorc:functions:intensity-histogram}
    &\delta(\mathcal{I}_{z},\mathcal{I}_{z'}) = 1 - HD(\mathcal{I}_{z},\mathcal{I}_{z'}) \\
    & HD(\mathcal{I}_{z},\mathcal{I}_{z'}) = \frac{1}{\sqrt{2}}\sqrt{\sum_\iota {\left(\sqrt{P(\mathcal{I}_{z})} - \sqrt{P(\mathcal{I}_{z'})}\right)}^2}
\end{split}
\end{align}
where $HD(\cdot)$ is the Hellinger distance and $P(\cdot)$ is the representation of an image as one-dimensional intensity histogram.

\subsubsection{Mutual Information (Mutual Info):}
\begin{align}
\begin{split}\label{eq:supp:sorc:functions:mutual-info}
    &\delta(\mathcal{I}_{z},\mathcal{I}_{z'}) = H(\mathcal{I}_{z}) + H(\mathcal{I}_{z'}) - H(\mathcal{I}_{z},\mathcal{I}_{z'}) \\
    &H(\mathcal{I}_{z}) = -\sum_{\iota} P(\mathcal{I}_{z}) log_e P(\mathcal{I}_{z}) \\
    &H(\mathcal{I}_{z},\mathcal{I}_{z'}) = -\sum_{\iota} P(\mathcal{I}_{z},\mathcal{I}_{z'}) log_e P(\mathcal{I}_{z},\mathcal{I}_{z'})
\end{split}
\end{align}
where $P(\cdot, \cdot)$ is a one-dimensional joint histogram.

\subsection{Clustering Methods}
\subsubsection{Hierarchical Clustering}
Hierarchical clustering is commonly used within the MSI community and in the entire realm of bioinformatics. One of its main advantages is that it allows to track each clustering step by inspecting the tree like result called dendrogram. For this study an agglomerative average linkage (UPGMA) procedure is applied \cite{sokal1958statistical}. 
The idea behind agglomerative average linkage hierarchical clustering is to initialize each data item (i.e.\ \mz-image) as a single cluster. Thereafter, the two clusters with the closest average distance are merged together. This procedure is applied iteratively until only one cluster remains or until the intended number of clusters is reached.
Without the use of approximation techniques, hierarchical clustering requires that the number of clusters be determined manually.
To avoid an arbitrary choice, a universally applicable approximation technique is used.
\begin{equation}\label{eq:thresholding}
	\varsigma = \mu(\Upsilon) + (c\ \sigma(\Upsilon))\\
\end{equation}
where $\Upsilon$ is the set of all distances resulting from all cluster merges, $\mu(\cdot)$ is the arithmetic mean, $\sigma(\cdot)$ is the standard variance and $c$ is a constant, which was set to one.

\subsubsection{Affinity Propagation}
The idea behind affinity propagation is to find data points that are suitable to be a representative of a number of other data points. To identify those points, messages are sent between pairs of samples until convergence. Each message belongs to one of two categories. First, the \emph{responsibility}, which describes how well one data item ($m$) is suited as a representative for another item ($n$) and second, the \emph{availability}, which describes how appropriate it would be for an item ($n$) to pick ($m$) as its representative \cite{frey2007clustering}. The method that detects the number of clusters automatically.
The applied implementation requires three parameters: the convergence iteration (set to 15), the maximum number of iterations (set to 200) and the damping factor (set to 0.5).

\subsubsection{Community Detection on Mass Channel Similarity Graphs}
The idea behind this method is to create a mass channel similarity graph from the set of all mass channel images and use community detection as clustering approach \cite{wullems2019detection}. The basic outline of this approach follows a three step procedure:
\begin{itemize}
    \item [1.] A pairwise similarity matrix is computed.
    \item [2.] A threshold is computed to transform the similarity matrix into an adjacency matrix, which is an alternative representation of the mass channel similarity graph.
    \item [3.] A community detection algorithm is applied.
\end{itemize}
The applied thresholding procedure is similar to Equation~\ref{eq:thresholding}, but in this case $\Upsilon$ equals the set of all pairwise similarity values of the similarity matrix.

\bibliographystyle{unsrt}
\bibliography{ms}  

\end{document}